# SATELLITE IMAGE CLASSIFICATION METHODS and LANDSAT 5TM BANDS

Jamshid Tamouk
Department of Computer Engineering
EMU University
Famagusta, North Cyprus
Jamshid.tamouk@cc.emu.edu.tr

Nasser Lotfi
Department of Computer Engineering
EMU University
Famagusta, North Cyprus
Nasser.lotfi@emu.edu.tr

Mina Farmanbar
Department of Computer Engineering
EMU University
Famagusta, North Cyprus
mina.farmanbar@emu.edu.tr

***Abstract*—This paper attempts to find the most accurate classification method among parallelepiped, minimum distance and chain methods. Moreover, this study also challenges to find the suitable combination of bands, which can lead to better results in case combinations of bands occur. After comparing these three methods, the chain method over perform the other methods with 79% overall accuracy. Hence, it is more accurate than minimum distance with 67% and parallelepiped with 65%. On the other hand, based on bands features, and also by combining several researchers' findings, a table was created which includes the main objects on the land and the suitable combination of the bands for accurately detecting of landcover objects. During this process, it was observed that band 4 (out of 7 bands of Landsat 5TM) is the band, which can be used for increasing the accuracy of the combined bands in detecting objects on the land.**

## I. INTRODUCTION

The focus of this paper is solely on some satellite image classification methods for land and also Landsat 5TM bands and the suitable combination of them to have higher accurate results during classification.

### A. *Classification methods*

The basic principle of the classification is classifying of images based on placing pixels with similar brightness value into the same group. It is done by selecting limited area or instance from images and then to assign the label (i.e. name) and color to that area. The images of the satellite (Landsat 5TM) which are used in this paper, has 7 bands for capturing the image of the earth. Each of these bands uses different wavelength for capturing the images, in a way that it causes to have 7 images from the same area but with different characteristics. The Landsat 5TM bands descriptions are shown in the Table 1:

TABLE 1: DESCRIPTION OF BANDS IN LANDSAT 5 TM [11]

| Band | Wavelength(μm) | Spectral | Resolution (m) |
|---|---|---|---|
| 1 | 0.45 – 0.52 | Blue-Green | 30 |
| 2 | 0.52 – 0.60 | Green | 30 |
| 3 | 0.63 – 0.69 | Red | 30 |
| 4 | 0.76 – 0.90 | Near IR | 30 |
| 5 | 1.55 – 1.75 | Mid-IR | 30 |
| 6 | 10.40 – 12.50 | Thermal IR | 120 |
| 7 | 2.08 – 2.35 | Mid-IR | 30 |

Methods of classification mainly follow two approaches, namely supervised and unsupervised classification. Supervised classification is the classification that needs to interact with user (i.e. training the system) who has knowledge (ground truth) about that area before image processing. However, in unsupervised classification, it is not necessary to have high knowledge about areas and it does not need to train the system. System starts grouping the pixels, which are similar in brightness value into unique clusters. Then after finishing clustering, the user will start to label each of the groups (classes).

The following two tables, which are resulted from some of the classification methods' comparison introduce by different researchers. According Table 2 (Hosseini et. al), overall accuracy of minimum distance (73.77%) is much better than parallelepiped method (34.27%) maximum likelihood method (with 85.83% overall accuracy) provides a higher accuracy than minimum distance.

TABLE 2: ACCURACY OF DIFFERENT CLASSIFICATION METHODS AND ALGORITHMS [6]

| Supervised Classification Method | Overall Accuracy % |
|---|---|
| Paralleleped | 34.27 |
| Minimum Distance | 73.77 |
| Maximum Likelihood | 85.83 |

Table 3 is the part of comparison table of classification methods, which is given by Todd [12]. According to this table, maximum likelihood method with 90.2% accuracy is a more accurate supervised classification in comparing with minimum distance (with 75.5% accurate) and parallelepiped (with 87.1% accurate).

TABLE 3: ESTIMATED ACCURACY AND PROCESSING TIME OF SOME CLASSIFICATION METHODS[12]

| **Method** | **Estimated process time** | **Accuracy %** |
|---|---|---|
| Maximum Likelihood | 18 min | 90.2 |
| Minimum Distance | 15 min | 75.5 |
| parallelepiped | 15 min | 87.1 |
| ISODATA | 2.25 min | 90.6 |

In comparing accuracy of maximum likelihood (supervised classification) with ISODATA method (unsupervised classification) according to the above table, we can see that their accuracies are approximately same. Here can see that opposite the previous table (table 1.2) parallelepiped is much accurate than minimum distance.
Based on the accuracy tables for supervised classification in the above tables (researches), maximum likelihood classification method is the most accurate method comparing with parallelepiped and minimum distance to mean methods and about parallelepiped and minimum distance there are different ideas. Some of researchers' finding like Hashemi el. al [4], Oruc el. al [9] and Todd [12] shows parallelepiped is more accurate than minimum distance but the others findings like Lim el. al [7] and Hosseini el. al [6] show the opposite of that.

### *B. 1.2 Suitable combination of the bands for land covers*

The other issue is about the way of combining the different bands of satellite for achieving a good result. In other words, for having more accurate result from classification of satellite images we should consider the selection of the suitable or correct combination of bands (according to the object or objects on the land, which we want to classify). According to the following formula, "[n! / (r! (n-r)!)]", 7 bands with three-band combination can have, 35 kinds of combination of bands. Below is the table of the combination of different bands and the corresponding "OIF Index" values, where OIF is used to show how useful is the combination of the bands, based on the correlations of them:
Table 4 below, which made by Wenbo el. al [13] for 20 selected bands combinations in to ascending order. In this table we can find that, OIF index of the combination of bands TM 3, 4, 5 (ETM+3, 4, 5) is the highest OIF index.
In this table from best 10 three-band combinations out of 20 three-band combinations, 8 of them include band 4 which gives highest number of occurrence in the combinations. It means 80% of 10 best three-band combinations (13**4**, 157, 357, 2**4**5, 1**4**5, 2**4**7, 1**4**7, **4**57, 3**4**7, and 3**4**5) have band **4** in their combination.

TABLE 4: OIF INDEX OF DIFFERENT BANDS COMBINATION BY ASCENDING ORDER [13]

| **Band combination** | **IOF index** | **Range Order** | **Band combination** | **IOF index** | **Range Order** |
|---|---|---|---|---|---|
| 123 | 12.832 | 1 | 13**4** | 22.605 | 11 |
| 127 | 16.739 | 2 | 157 | 22.724 | 12 |
| 124 | 17.229 | 3 | 357 | 22.840 | 13 |
| 237 | 18.043 | 4 | 2**4**5 | 23.918 | 14 |
| 125 | 18.359 | 5 | 1**4**5 | 24.316 | 15 |
| 137 | 19.160 | 6 | 2**4**7 | 24.858 | 16 |
| 135 | 19.693 | 7 | 1**4**7 | 25.724 | 17 |
| 235 | 20.596 | 8 | **4**57 | 27.442 | 18 |
| 257 | 21.314 | 9 | 3**4**7 | 29.209 | 19 |
| 234 | 22.169 | 10 | 3**4**5 | 29.230 | 20 |

In Table 5, which made by Hobson [5] all possible combination of bands (35 set of combined bands) in three-band combination has been examined which the best combination of the bands belongs to the bands 4, 5, 6 with 57.3673 OIF. According to the below table it can be understood that from the best 10 tree-band combination out of 35, eight of these combination include band 4 in their combination. It means 80% of 10 best three-band combinations (145, 457, 167, 246, 347,146, 346, 356, 467, and 456) have band 4 in their combination.

TABLE 5: OIF OF 35 COMBINED BANDS [5]

| **Bans Combination** | **OIF** | **Bans Combination** | **OIF** |
|---|---|---|---|
| 123 | 12.6385 | 237 | 16.1890 |
| 124 | 18.9822 | 245 | 27.8149 |
| 125 | 20.2492 | 2**4**6 | 35.9016 |
| 126 | 15.6910 | 247 | 23.4820 |
| 127 | 15.0502 | 256 | 28.8717 |
| 134 | 22.7656 | 257 | 22.9567 |
| 135 | 22.8736 | 267 | 20.3396 |
| 136 | 18.8254 | 345 | 31.3270 |
| 137 | 17.5892 | 3**4**6 | 42.0967 |
| 1**4**5 | 31.9984 | 3**4**7 | 39.9820 |
| 1**4**6 | 40.1405 | 356 | 44.0859 |
| 147 | 26.9859 | 357 | 30.0630 |
| 156 | 29.5769 | 367 | 24.4388 |
| 157 | 24.5532 | **4**56 | 57.3673 |
| 167 | 35.7702 | **4**57 | 33.7486 |
| 234 | 19.5080 | **4**67 | 46.6954 |
| 235 | 21.0872 | 567 | 31.8615 |

According to other researchers' findings mentioned below and after finding the best bands combination for classification of land covers, now we want to see which combination of bands are suitable for each of the main objects (such as water, vegetation, soil, snow and ice, sand and so on ) on the land.
In order to recognize the water on land surface or distinguish between land and water (coastal and see) a combination of band2 and band5 can be used as below:
The ratio band2/band5 is greater than one for water and less than one for land in large areas of coastal zone. With this method, water and land can be separated directly.

Another method is to use the band ratio between band 4 and band 2 [1]. So, if the result of band2/band5 become greater than one (band2/band5>1) then the object will be water; otherwise, it will be land or any other objects. Also in the following combination of bands is used for calculating the water index and classifying the images based on water and non-water by Hosseini el. al [6]:

Water Index = (Band 1 + Band 2 + Band 3) / (Band 4 + Band 5 + Band 7) [4]. That is the ratio of visible spectrum bands to be reflected by infrared bands.

The other important object on the land is vegetation or the area with vegetative cover. According to the Shan long el. al [10] finding, the best combination of bands for detecting or recognizing the vegetation is combination of band4 and band3.

NDVI = (Band 4 - Band 3) / (Band 4 + Band 3) [10]

This combination of bands 3 and 4 is called "normalized difference vegetative index" (NDVI). In above formula the bands which are used are NIR = Reflectance in Near Infrared Band (band 4) and RED = Reflectance in the RED Band (band 3). It is often used in detection of small differences between vegetation classes and sometimes used for distinguishing vegetative area from other areas or objects. In these bands, vegetation and soil contrast is at its maximum. It means soil and vegetation can be easily differentiated from each other.

The NDVI (Normalized Difference Vegetative Index) values in the range of -1.0 to 1.0, where Vegetated areas will typically have values greater than zero and Negative values indicate non-vegetated surface features such as water, barren, ice, snow, or clouds [10].

So, if NDVI is greater than zero (if NDVI > 0) then that area is the vegetative cover area; otherwise, it is land (other objects). Following is the other vegetative index formulas, which are discussed by Muzein [8]:

Corrected Naturalized Differential Vegetation Index

$$\left(\frac{Band\,4 - Band\,3}{Band\,4 + Band\,3}\right) \times \left[1 - \frac{Band\,5 - (Band\,5)_{MIN}}{(Band\,5)_{MAX} - (Band\,5)_{MIN}}\right]$$

Percent Vegetation Cover:

$$(\text{Standardized} \quad \text{NVDI})^2$$

Simple Ratio $\left(\frac{Band\,4}{Band\,3}\right)$

Reduced Simple Ratio:

$$\left(\frac{Band\,4}{Band\,3}\right) \times \left[1 - \frac{Band\,5 - (Band\,5)_{MIN}}{(Band\,5)_{MAX} - (Band\,5)_{MIN}}\right]$$

Soil Adjusted Vegetation Index: Minimizes the secondary backscattering effect of canopy-transmitted soil background reflect radiation. It describes both vegetation cover and soil background.

$$\left(\frac{Band\,4 - Band\,3}{Band\,4 + Band\,3 + 1}\right) \times (1 + L)$$

L depends on Land cover, but 0.5 is a suggested value for many land cover conditions

According to Hashemi el. al [4] research finding for discriminating or distinguishing between salt and sodium soils, use of Landsat 5TM bands (2,4,6) and 7 are more accurate than other combination of the bands.

Also according to the Hashemi el. al [4] finding the ratio spectral (TM3 - TM4) / (TM2 - TM4), show the best correlation with the Soil EC data. It means that by combination of the band2, band3 and band4 we can distinguish different kinds of soil from each other and other objects.

The other objects on surface of the land, are snow and ice (i.e. glacier area) which based on the Todd research finding [12] can be detected or recognized by using the combination of bands 4 and 5 (ice or snow index = band4 / band5), bands 3 and 5 (ice or snow index = band3 / band5), or bands 3, 4 and 5.

"Glacier extent mapping from satellite data has been the focus of many recent research papers. Bayr et al. (1994) used a threshold of a ratio image of TM 4 to TM 5 bands to delineate glacier area, while Rott (1994) used a threshold of a TM 3 to TM 5 ratio image. Paul (2000) found that the TM 4 to TM 5 ratio technique yielded the best results for glacier mapping on Gries Glacier, especially in regions with low insolation" [12].

"Also using the visible-red channel (TM 3) and two of the infrared channels (TM 4 and 5) allowed for an excellent distinction of the ice cap. This is because snow and ice have very high spectral reflectance in the visible-red (RED) and the near-infrared (NIR) wavelength regions and very low spectral reflectance's in the middle-infrared (MIR) wavelength region" [12].

Based on the above discussion, the best combination for distinction of ice and snow is combination of bands 3, 4 and 5.

Below there are some other combinations of the bands which are, the findings of several researchers. Based on the result of the below researches, combination of bands 3, 4, 7 is good for detecting the water boundary or costal, soil moisture and iron compounds. Bands 4,3, 2 are used for vegetation and crop analysis, bands 4, 5, 3 for soil moisture and vegetation analysis, bands 3, 2, 1 for landcover and underwater features, bands 7, 4, 3 and bands 7, 4 , 2 for change detection, soil type and vegetation stress. [2, 3 and 7]

## II. Problem Definition

Researchers depending on their application should be able to choose suitable method for classification of their images. Obviously, it is difficult for the researchers who are new in this field. On the other hand, most of the researchers who use Landsat images in their research need to know more about the Landsat 5TM bands and usage of each band. In addition, they should know which bands and which combination of the bands are good for detecting different kinds of objects on the land. Therefore, they will be able to get a good result if they are able to choose suitable methods and also suitable bands or combination of the bands for their research.

## III. PROPOSED METHOD

By using Landsat 5TM images as the input data (captured from north of Iran) and with training the system by the information collected about that area (ground truth) for supervised methods and without training the system for unsupervised method achieved to the ability to classify satellite images and also calculate the accuracy of each methods. Then the classification methods (supervised and unsupervised classification) are compared based on the determined accuracy. On the other hand, Table 8 is created based on the bands' features to illustrate the main objects on the land and the suitable combinations of bands for recognizing them. According to this table and by checking the other research findings on bands and band combinations, it becomes possible to find the most effective band among 7 bands of Landsat 5TM.

## IV. PERFORMANCE ANALYSIS

Accuracy is calculated by dividing the number of object/class's pixels correctly classified; over the total number of pixel belong to that object/class. In other word, Accuracy= number of pixels assigned to the correct class / number of pixels that actually belong to that class or object. If this calculation is done for all of objects/classes together, the result will call Overall accuracy. Table 6 shows the confusion matrix of minimum distance method, fund for this paper (It is a table that shows the correct and incorrect number of assigned objects to each class. It is used for computation of accuracy).

TABLE 6: ERROR MATRIX TABLE FOR MINIMUM DISTANCE TO MEAN

| Class type form classification | Class types determined from reference source | | | | | |
|---|---|---|---|---|---|---|
| | #Plots | Water | Forest | Grassland Agriculture | Mountain and soil | Totals |
| | Water | 39 | 0 | 8 | 4 | 52 |
| | Forest | 0 | 32 | 12 | 4 | 48 |
| | Grassland & Agriculture | 4 | 8 | 33 | 8 | 52 |
| | Mountain | 0 | 8 | 8 | 24 | 40 |
| No ground truth pixels | | 43 | 48 | 61 | 40 | 192 |

## V. RESULT AND DISCUSSION

Table 7 is the accuracy table which is created based on the parallelepiped, minimum distance and chain methods error matrix tables gained for this paper. According to this table, the chain method (with 79% accuracy) has highest accuracy in comparing with two others mentioned methods in this paper. Also it is found that minimum distance (with 67% accuracy) has higher accuracy than parallelepiped (with 65% accuracy). Some of researchers' findings about accuracy of parallelepiped and minimum distance approximately are similar to my research finding such as Table 2 but some others are exactly opposite such as Table 3. The reasons for this could be one of the following reasons: lack of enough data for training or testing, samples distribution, difficulty with selecting sufficient training data for supervised methods or the insufficient skill of the trainers.

TABLE 7: USER, PRODUCER AND OVERALL ACCURACY OF PARALLELEPIPED, MINIMUM DISTANCE AND CHAIN METHOD

| Methods | Objects | Water | Forest | Agriculture | Mountain |
|---|---|---|---|---|---|
| **Paralleled piped** | **User accuracy** | 68% | 72% | 50% | 71% |
| | **Producer accuracy** | 92% | 62% | 54% | 50% |
| | **Overall accuracy** | [(43+31+27+23)/192]*100=**65%** | | | |
| **Minimum Distance** | **User accuracy** | 77% | 67% | 62% | 60% |
| | **Producer accuracy** | 91% | 67% | 53% | 60% |
| | **Overall accuracy** | [(39+32+33+24)/192]*100=**67%** | | | |
| **Chain Method** | **User accuracy** | 86% | 77% | 73% | 80% |
| | **Producer accuracy** | 92% | 67% | 73% | 89% |
| | **Overall accuracy** | [(45+41+32+34)/192]*100=**79%** | | | |

Below is the table of some important objects and the suitable combination of the bands for detecting those objects based on the literature given in section B.

TABLE 8: COMBINATION OF BANDS BASED ON THE TYPE OF OBJECT

| | Bands | | | | | | | |
|---|---|---|---|---|---|---|---|---|
| **Objects** | **1** | **2** | **3** | **4** | **5** | **6** | **7** | **Combination of the bands & Conditions** |
| **Water** | | * | | | * | | | TM 2/TM 5>1 |
| | | * | | * | | | | TM 2/TM 4>1 |
| | * | * | * | * | * | | * | Index=(TM 1+TM 2+TM 3)/(TM 4+TM 5+TM 7) |
| | | * | * | * | | | | Water appears dark |
| **Coastal** | | | * | * | | | * | |
| **Vegetation** | | | * | * | | | | (TM 4-TM 3)/(TM 4+TM 3)>0 |
| | | * | * | * | | | | False color infrared |
| | | | * | * | * | | | Vegetation conditions |
| **Snow and Ice** | | | | * | * | | | Index=TM 4/TM 5 |
| | | | * | | * | | | Index=TM 3/TM 5 |
| | | | * | * | * | | | |
| **Soil type** | | | * | * | | | * | |
| | | * | | * | | | * | |
| **Salt and Sodium Soil** | | * | | * | | * | * | |
| | | * | * | * | | | | (TM 3-TM 4)/(TM 2-TM4) |
| **Iron Compounds** | | | * | * | | | * | Such as ilmenite |
| **Soil moisture differences** | | | * | * | | | * | Best combination |
| | | | * | * | * | | | |
| **Change detection, disturbed soils vegetation stress** | | | * | * | | | * | |
| | | * | | * | | | * | |

Table 8 shows the important object on the land and the best combination of the bands for detecting them. It is important to know that which combination of the bands can detects which kind of object on the land with more accuracy. In Table 8, by carefully looking at this table, we can find that band 4 is used in all of the objects in most of the combined bands (around 90% of the combined bands). It can also be observed in Table 4 that from the best 10 three-band combinations out of 20, eight of them (i.e. 80% of ten best three-band combinations) include band 4 in their combinations (134, 157, 357, 245, 145, 247, 147, 457, 347, and 345). The same result is approximately shown in Table 5. In this table from the best 10 three-band combinations out of 35, eight of these combinations include band 4 in their combinations (145, 457, 167, 246, 347,146, 346, 356, 467, and 456) which is 80% of 10 best three-band combinations. Therefore, these findings confirm that band 4 is the most useful band to increase the accuracy of combined bands in detecting the objects on the land.

## VI. Conclusion

According to this paper, the proposed chain method with 79% accuracy is more accurate than the other two compared methods. In addition, Table 8, which identifies the suitable band combinations for each of the main objects on the land, was created. Finally, after analyzing the findings of this paper and some other researchers, similarly it is concluded that, having band 4 (Near Infrared) in the combinations of the bands can improve the accuracy of detection and classification of the objects noticeably.

## References

[1] Alesheikh, A.A., Ghorbanali, A and Nouri, N. (2007). *Coastline change detection using remote sensing.* ISSN: 1735-1472, IRSEN, CEERS, IAU.

[2] CURRENT SCIENCE. (2008). *Discovery of heavy mineral-rich sand dunes along the Orissa–Bengal coast of India using remote sensing techniques*. CURRENT SCIENCE, VOL. 94, NO. 8, 25.

[3] Davis, A and Allen, J. *Landsat 7 at NASA Goddard Space Flight Center*. [online] viewed (December 2012) http://landsat.gsfc.nasa.gov

[4] Hashemi, S.S., Baghernejad, M., Pakparvar, M and Emadi, M. (2005). *GIS classification assessment for mapping soils by satellite images.* Soil Science Dept, College of Agriculture. Shiraz University. Shiraz, Iran.

[5] Hobson, V. R. (2003). *Characterization of Craters of the Moon Lava Field using Landsat TM data*. FinalReport FRWS 6750 – Applied Remote Sensing.

[6] Hosseini, S.Z., Khajeddin, S.J and Azarnivand, H. (2008). *Application of ETM+data for estimating rangelands cover percentage*. College of Natural Resources & Desert Studies', University of Yazd, Yazd, Iran.

[7] Lim, H.S., MatJafri, M.Z and Abdullah, K. (2002). *Evaluation of conventional digital camera scenes for thematic information extraction.* School of Physics Universiti Sains Malaysia, Penang, Malaysia

[8] Muzein, B.S. (2006). *Remote Sensing & GIS for Land Cover/ Land Use Change Detection and Analysis in the Semi-Natural Ecosystems and Agriculture Landscapes of the Central Ethiopian Rift Valley*. PHD research in Technische Universität Dresden.

[9] Oruc, M., Marangoz, A. M and Buyuksalih, G. (2004). *Comparison of pixel based and object oriented classification approaches using Landsat 7ETM spectral bands.*

[10] Shan-long, L., Xiao-hua, S and Le-jun. Z. (2006). *Land cover change in Ningbo and its surrounding area of Zhejiang Province.* Journal of Zhejiang University SCIENCE A ISSN 1009-3095 (Print); ISSN 1862-1775 www.zju.edu.cn/jzus.

[11] Short, N. M. *Technical and historical perspectives of remote sensing.* [online] viewed (December 2012), Available at: http://ltpwww.gsfc.nasa.gov/IAS/handbook/handbook_toc.html.

[12] Todd H. A. (2002). *Evaluation of remote sensing techniques for ice area classification applied to the tropical quelccaya icecap*, Peru, Polar Geography, 2002, 26, No. 3.

[13] Wenbo,W., Jing, Y and Tingjun, K. (2008). *Study of remote sensing image fusion and its application in image classification*, *The International Archives of the Photogrammetry.* Remote Sensing and Spatial Information Sciences. Vol. XXXVII.